\documentclass{article}
\usepackage{ijcai25}

\usepackage{times}
\usepackage{soul}
\usepackage{url}
\usepackage[hidelinks]{hyperref}
\usepackage[utf8]{inputenc}
\usepackage[small]{caption}
\usepackage{graphicx}
\usepackage{amsmath}
\usepackage{amsthm}
\usepackage{booktabs}
\usepackage{algorithm}
\usepackage{algorithmic}
\usepackage[switch]{lineno}

\usepackage[utf8]{inputenc}
\usepackage{geometry} % For setting page margins
\usepackage{tabularx} % Required for tabularx environment
\usepackage{booktabs} % For professional-looking table rules (recommended over \hline)
\usepackage{caption}  % For \captionof if needed, or if using subfigures
\usepackage{amsmath}  % For general math
\usepackage{amssymb}  % For math symbols
\usepackage{amsfonts} % For math fonts
\usepackage{hyperref} % For clickable citations
\usepackage{cleveref} % For \cref command

\newcolumntype{Y}{>{\hsize=.5\hsize}X} % Half the normal X width (relative to total X-width)
\newcolumntype{Z}{>{\hsize=2\hsize}X} % Twice the normal X width (relative to total X-width)

\title{Early Prediction of Multiple Sclerosis Disability Progression via Multimodal Foundation Model Benchmarks}

\author{
Maxime Usdin$^1$
\and
Lito Kriara$^2$\and
Licinio Craveiro$^2$\\
\affiliations
$^1$Genentech, Inc, South San Francisco, CA, USA\\
$^2$F. Hoffmann-La Roche Ltd., Basel, Switzerland\\
\emails
usdin.maxime@gene.com,
\{lito.kriara, licinio.craveiro\}@roche.com
}
\begin{document}

\maketitle

\begin{abstract}
Early multiple sclerosis (MS) disability progression prediction is challenging due to disease heterogeneity. This work predicts 48- and 72-week disability using sparse baseline clinical data and 12 weeks of daily digital Floodlight data from the CONSONANCE clinical trial. We employed state-of-the-art tabular and time-series foundation models (FMs), a custom multimodal attention-based transformer, and machine learning methods. Despite the difficulty of early prediction (AUROC 0.63), integrating digital data via advanced models improved performance over clinical data alone. A transformer model using unimodal embeddings from the Moment FM yielded the best result, but our multimodal transformer consistently outperformed its unimodal counterpart, confirming the advantages of combining clinical with digital data. Our findings demonstrate the promise of FMs and multimodal approaches to extract predictive signals from complex and diverse clinical and digital life sciences data (e.g., imaging, omics), enabling more accurate prognostics for MS and potentially other complex diseases.
\end{abstract}

\section{Introduction}\label{intro}

Early prediction of multiple sclerosis (MS) disability progression is vital for personalized care, but is hindered by measurement limitations and disease heterogeneity \cite{oh_use_2024}. Numerous studies highlight this prognostic challenge, attempting to predict risk using diverse machine learning approaches on varied data (from baseline clinical, imaging and transcriptomic data \cite{eshaghi_predicting_2022,yousef_predicting_2024,gurevich_machine_2025}, to longitudinal clinical registries \cite{de_brouwer_machine-learning-based_2024} and explorations of digital measure correlates \cite{graves_assessment_2023}). Despite these varied efforts and the potential of advanced computational methods, achieving high predictive performance for early MS disability progression remains an unmet clinical and research need. These efforts, utilizing inputs ranging from purely baseline assessments to extensive clinical histories, underscore the difficulty of early prediction despite the potential of advanced computational methods.

Foundation models (FMs) show promise in single modalities \cite{bommasani_opportunities_2022}, and Large Language Models (LLMs) are being applied to diverse data types. However, few studies have systematically developed unified architectures for MS progression that combine sparse clinical data with dense, potentially noisy, digital biomarkers across different time scales \cite{dillenseger_digital_2021,simon_future_2024,wenk_building_2024}. Our approach conceptually extends pre-trained multimodal large language models (LLMs) by treating sparse clinical assessments, dense digital biomarker streams, and other relevant structured data as varied tokens within a unified sequence, adaptable to sequence modeling.

This work addresses these gaps by comprehensively assessing early MS disability prediction using three months of observational data. We evaluated tabular and time-series FMs, and our novel multimodal transformer architecture (which fuses sparse clinical features with dense Floodlight time-series data inspired by the canonical transformer model \cite{vaswani_attention_2023}), against state-of-the-art methods. Our findings reveal the potential of foundation models and multimodal data integration for early prediction, guiding the development of future clinical multimodal models.

\section{Methods}

\subsection{Data}

Data were sourced from 811 participants enrolled in the CONSONANCE Phase 3b clinical trial (NCT03523858) \cite{comi_multicentre_2022}. This dataset uniquely combines infrequent clinical assessments with frequent, patient-generated digital biomarker data collected via the Floodlight smartphone technology \cite{montalban_smartphone_2022}. Clinical data included baseline demographics, disease characteristics, and longitudinal assessments of neurological function and performance such as the Expanded Disability Status Scale (EDSS), Timed 25-Foot Walk (T25FWT), 9-Hole Peg Test (9HPT), and other standard clinical outcomes, typically collected every 24 weeks \cite{kurtzke_rating_1983,cutter_1999}. Digital biomarker data consisted of CONSONANCE's initial 12 weeks of daily time-series measurements of Floodlight tests designed to measure hand function and gait (i.e., Draw a Shape \cite{graves_preliminary_2023}, Pinching Test \cite{graves_assessment_2023}, U-Turn Test \cite{cheng_2021}, and the 2-Minute Walk Test \cite{montalban_smartphone_2022}). The target variable for prediction was 24-Week composite Confirmed Disability Progression (24W cCDP), a standard composite endpoint in MS trials reflecting clinically meaningful worsening. The distribution of patient data used is summarized in Table \ref{tab1-dem}. A complete list of the features used is shown in Table \ref{tab:data_overview}. 

\begin{table}[t]
% \vskip 0.15in
\begin{center}
\begin{small}
\begin{sc}
\begin{tabular}{lcccr}
\toprule
Variable & Week 48 & Week 72 \\
\midrule
Participant number & 406 & 394  \\
 with clinical data, n & &  \\
Age at Baseline & 48.8 (9.3) & 48.7 (9.4) \\
 (years), mean (SD) & &  \\
Female, n (\%)    & 219 (53.9) & 210 (53.3) \\
EDSS, mean (SD)   & 4.8 (1.4) & 4.9 (1.5) \\
T25FWT, mean (SD) & 16.2 (26.3) & 20.5 (35.1) \\
9HPT, mean (SD)  & 35.9 (29.1) & 36.1 (27.3) \\
cCDP events, n (\%)  & 108 (24) & 155 (35) \\
% cCDP events in test, n (\%) & 29 (27) & 44 (42) \\
\bottomrule
\end{tabular}
\end{sc}
\end{small}
\caption{Demographic information of the CONSONANCE dataset.}
\label{tab1-dem}
\end{center}
\vskip -0.2in
\end{table}

\subsection{Preprocessing}\label{preproc}

\begin{figure}[t]
% \vskip 0.2in
\begin{center}
% \centerline{\includegraphics[width=\columnwidth]{figures/Figure1_cropped.jpg}}
\centerline{\includegraphics[width=\columnwidth]{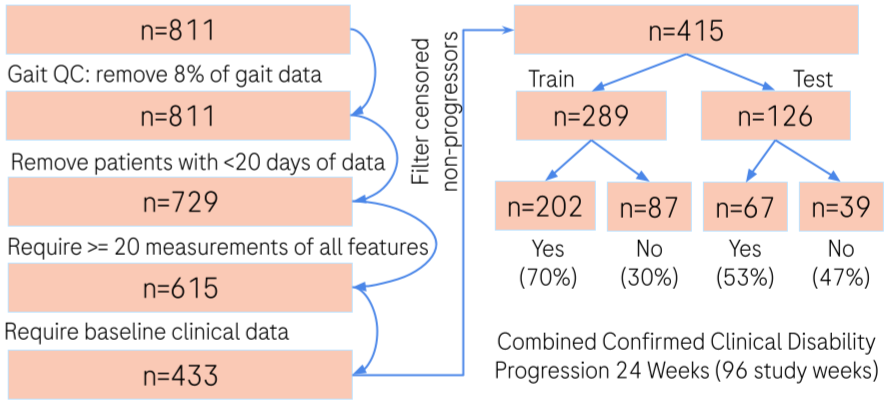}}
\caption{Study population selection. Gait features were filtered using pre-determined quality control flags in the Floodlight app. %Of the original 811 patients, 729 were selected with at least 20 unique days of data in the first 12 weeks of the study. Further filtering with a minimum data requirement of 20 measurements for each Floodlight feature resulted in 615 patients. Requiring complete baseline clinical data reduced the population to 433 patients. From there, patients who dropped off the study before 72 weeks without experiencing a progression event (censored non-progressors) were removed, yielding the final dataset of 415 patients. These were split into a train set (289 patients) and test set (126 patients) using a pre-defined date cutoff from the Floodlight data.
}
\label{fig-population}
\end{center}
\vskip -0.2in
\end{figure}

Clinical data were one-hot encoding for categorical variables and standard scaling was applied to numerical features. Rigorous cohort filtering was performed to ensure baseline data availability and sufficient longitudinal density to model progression dynamics, resulting in the final analysis cohort of 415 out of originally 811 study participants (\cref{fig-population}). A fixed 80/20\% train/test split was established based on a pre-defined data cut-off date to ensure temporal validity, preventing data leakage into the training set.

\subsection{Feature Extraction and Gaussian Process Interpolation}

To capture temporal dynamics from the Floodlight time series within the 12-week input window, feature engineering approaches were employed, including calculating windowed statistics (mean, variance, trend coefficients) and utilizing the \textit{tsfresh} \cite{christ_time_2018} library for automated extraction of a comprehensive set of time-series characteristics. Feature selection methods, including ANOVA F-test and Recursive Feature Elimination (RFE), were used to reduce dimensionality and select informative features.

\begin{figure}[t] 
    \centering 
    \includegraphics[width=0.8\columnwidth]{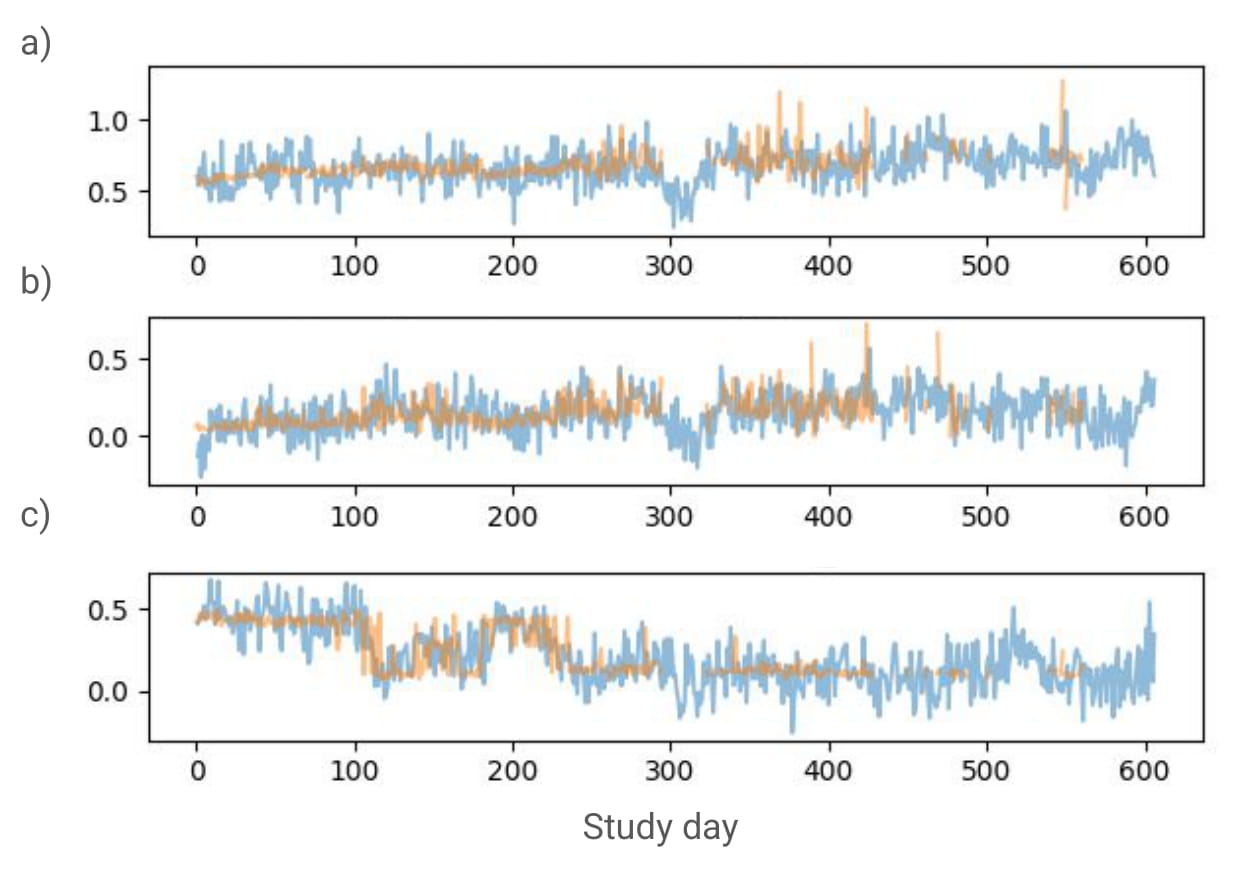}
    % \refstepcounter{figure}
    % \renewcommand{\thefigure}{S1}
    \vskip -0.1in
    \caption{Illustrative examples of Floodlight digital biomarker time-series data from a single patient, augmented using patient-specific Gaussian Processes (GPs). The figure displays three trajectories over the study duration: (a) step duration (median), (b) step duration (interquartile range, IQR), and (c) step length (median). Each panel shows the original, more variable observed data points (orange) for the respective metric, alongside a sampled trajectory from a fitted GP model (blue). %These visualizations demonstrate how GPs capture individual temporal patterns for different gait parameters and are used to interpolate missing values and generate complete, synthetic trajectories for subsequent model training and data augmentation.
    }
    \label{fig-gp}
\end{figure}

To address missing data within the dense but potentially incomplete daily Floodlight measurements, we employed patient-specific Gaussian Processes (GPs). GPs offer a robust probabilistic framework for interpolating and modeling temporal trajectories from such sparse and noisy data. We fitted these patient-specific GPs to model the evolution of individual Floodlight features. The resulting models were then used to generate complete trajectories for downstream model training: observed Floodlight measurements were preserved, while any missing time-points in these original trajectories, along with entirely synthetic trajectories for data augmentation, were instantiated by sampling from the posterior predictive distribution of the fitted GPs. Illustrative examples of observed data points and their corresponding GP-imputed trajectories are provided in \cref{fig-gp}. 

%Full mathematical details of the GP implementation, including kernel selection and hyperparameter optimization, are provided in the Supplemental Material. %\cref{supplement}.

To implement the GP methodology, we utilized a composite kernel combining a Radial Basis Function (RBF) for smoothness and a White Noise kernel to account for measurement noise. The covariance between two time points $t_i$ and $t_j$ is given by:

\begin{equation*}
    k(t_i, t_j) = \sigma_c^2 \exp \left( -\frac{(t_i - t_j)^2}{2l^2} \right) + \sigma_n^2 \delta_{ij}
\end{equation*}

where $\delta_{ij}$ is the Kronecker delta, which is 1 if $i=j$ and 0 otherwise. The hyperparameters $\theta = \{\sigma_c^2, l, \sigma_n^2\}$ represent output variance, characteristic length-scale, and noise level, respectively.

Prior to optimization, initial values and bounds for $\theta$ were estimated heuristically for each patient-feature pair based on data characteristics (overall feature variance for $\sigma_c^2$, time intervals for $l$, and linear fit residuals for $\sigma_n^2$). Within these bounds, the hyperparameters were then optimized by maximizing the log marginal likelihood:

\begin{equation*}
    \log p(\mathbf{y} | \mathbf{t}, \theta) = -\frac{1}{2} \mathbf{y}^T K^{-1} \mathbf{y} - \frac{1}{2} \log |K| - \frac{n}{2} \log 2\pi
\end{equation*}

Here, $\mathbf{y}$ is the vector of observed values, $\mathbf{t}$ is the vector of corresponding time points, and $K$ is the covariance matrix where $K_{ij} = k(t_i, t_j)$. The posterior predictive distribution at a new time point $t_*$ given observed data $(t, y)$ is:

\begin{equation*}
    p(f_* | t_*, \mathbf{t}, \mathbf{y}, \theta) = \mathcal{N}(\mathbf{k}_*^T K^{-1} \mathbf{y}, k(t_*, t_*) - \mathbf{k}_*^T K^{-1} \mathbf{k}_*)
\end{equation*}

where $\mathbf{k}_*$ is the vector of covariances between the new time point $t_*$ and the observed time points $\mathbf{t}$, i.e., $\mathbf{k}_* = [k(t_*, t_1), k(t_*, t_2), ..., k(t_*, t_n)]^T$. This posterior distribution allows us to sample synthetic data $f_*$ conditioned on our observed data and optimized hyperparameters $\theta$.

\subsection{Model Architectures}
A range of unimodal and multimodal models was evaluated:

\textit{Machine Learning:} AutoGluon \cite{erickson_autogluon-tabular_2020}, an automated machine learning framework, was used to establish robust baselines on tabular feature sets derived from clinical data and/or aggregated longitudinal Floodlight features. We also trained XGBoost, Logistic Regression, and Random Forest models as additional competitive baselines.

\textit{Time-Series Modeling:} Clinical and Floodlight time-series features data were concatenated and fed into a comprehensive suite of state-of-the-art time-series classifiers from the \textit{sktime} library \cite{loning_sktime_2019}. We utilized a diverse set of time-series classification models, encompassing kernel-based (RocketClassifier), interval-based (CanonicalIntervalForest, DrCIF), distance-based (KNeighborsTimeSeriesClassifier), hybrid (HIVECOTEV2), deep learning (MVTSTransformerClassifier, CNNClassifier, ResNetClassifier), dictionary-based (MUSE), and feature-based (FreshPRINCE) approaches, with model selection facilitated by grid search cross-validation on the training set.

\textit{Foundation Model Approaches (Unimodal):} 
\begin{itemize}
    \item TabPFN \cite{hollmann_accurate_2025}:A transformer-based FM optimized for small tabular datasets, applied here to clinical data combined with engineered Floodlight features. Its inclusion aims to test the applicability of tabular FMs to this mixed-feature setting.
    \item Moment+Transformer: Precomputed time-series embeddings were generated using the publicly available MOMENT FM \cite{goswami_moment_2024}. The embeddings, capturing rich temporal patterns from the Floodlight data, were fed as input sequences into a standard transformer encoder architecture for classification.
    \item GP-Transformer: A standard transformer encoder architecture applied directly to the GP-augmented Floodlight time-series data.

\end{itemize}

\begin{figure}[t]
% \vskip 0.2in
\begin{center}
\centerline{\includegraphics[width=\columnwidth]{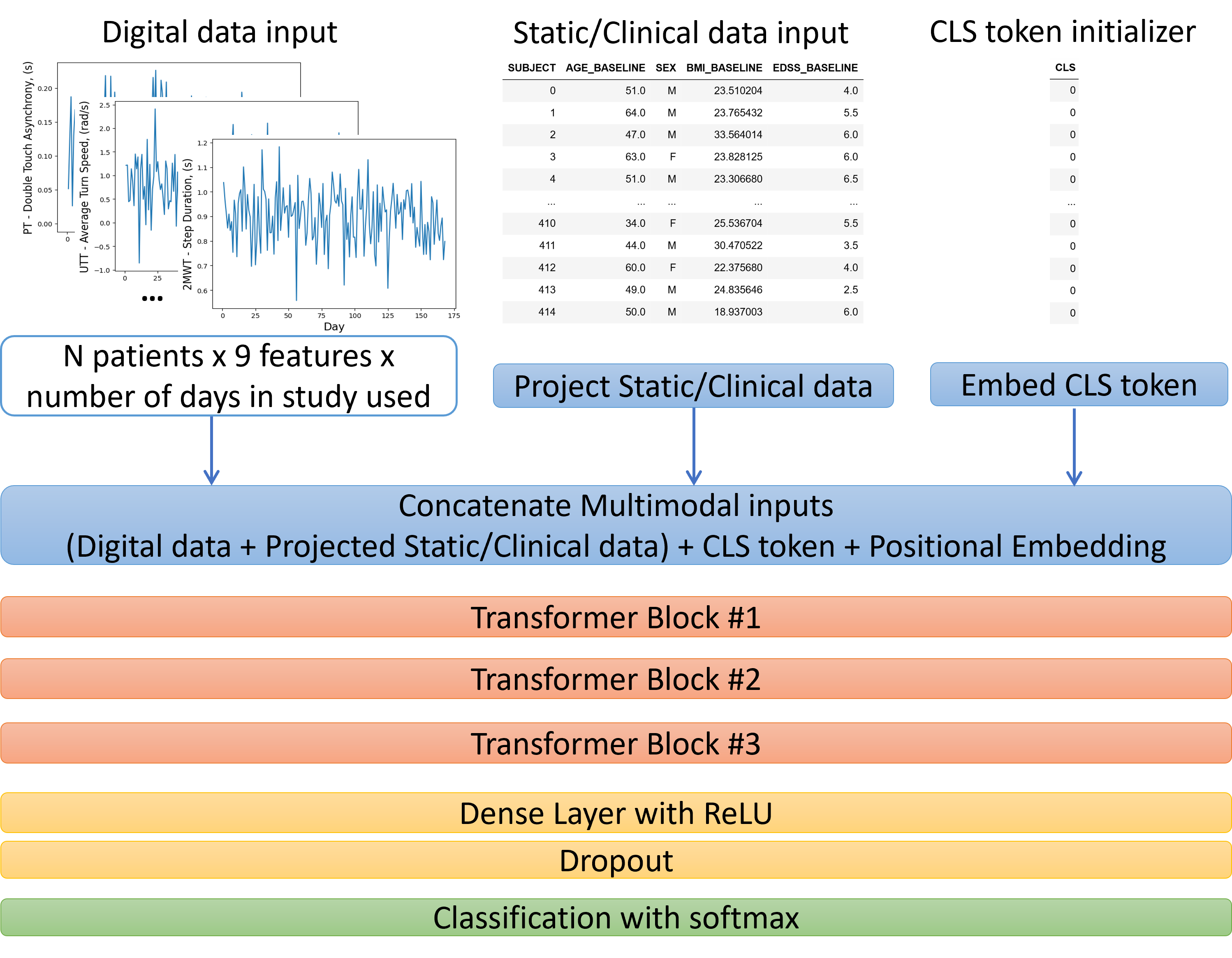}}
\caption{Schematic of the custom multimodal transformer, illustrating the integration of GP-processed time series, projected static/clinical features, and a CLS token prior to processing through transformer blocks and an MLP classification head.}
\label{fig-trans-arch}
\end{center}
\vskip -0.2in
\end{figure}

\textit{Custom Multimodal Transformer:} An attention-based multimodal transformer model (\cref{fig-trans-arch}) was specifically engineered to seamlessly integrate heterogeneous inputs. It takes in three distinct data streams: Gaussian-Process augmented Floodlight data, projected static/clinical features, and a classification (CLS) token. These are concatenated along the sequence dimension, enhanced with positional embeddings, and then processed through standard transformer blocks. To ensure compatibility, the static features are projected to match the time-series embedding dimension using a feed-forward network. Finally, the CLS token's representation, which aggregates information through self-attention across the entire fused sequence, is passed through final MLP layers for classification.

% This architecture (\cref{fig-trans-arch}) was designed specifically to integrate heterogeneous inputs. It utilized three input streams (GP augmented Floodlight, projected static/clinical features, and a classification (CLS) token), concatenated along the sequence dimension, augmented with positional embeddings, and processed through standard transformer blocks. The static features were projected to match the time-series embedding dimension using a feed-forward network. The CLS token representation, aggregated through self-attention across the fused sequence, was passed through final MLP layers for classification.

\subsection{Training and Evaluation}
Models were trained on the 80\% training set, with 5-fold CV for hyperparameter optimization for machine learning models. Final model performance was evaluated on the held-out 20\% test set (\cref{fig-population}) using the AUROC as the primary metric, which is well-suited for imbalanced classification tasks such as early disability progression prediction.

\section{Results}\label{res}

\begin{figure}[t] 
    \centering 
    \begin{minipage}{\columnwidth} 
        \centering % Centers the image within the minipage
        \includegraphics[width=\linewidth]{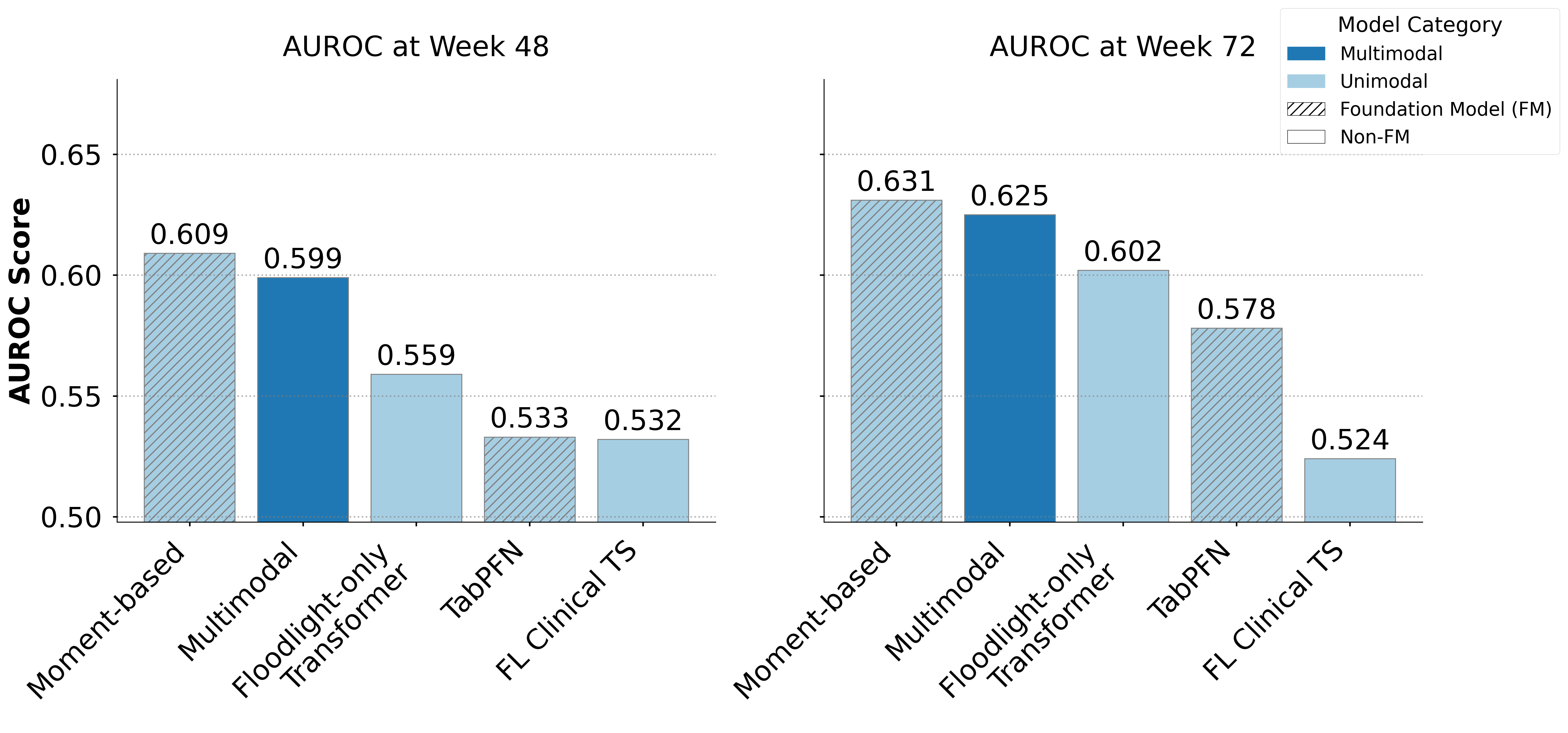}
    \end{minipage}% 
    \caption{Model performance (AUROC) for predicting 48-week (left) and 72-week (right) cCDP using 12 weeks of input data with Moment-based, embeddings derived from the Moment time-series foundation model; Multimodal, our multimodal transformer; Floodlight-only transformer, our transformer model trained on Floodlight data only; TabPFN, a tabular foundation model trained on clinical and derived Floodlight features; and FL Clinical TS, which denotes the optimal machine learning model performance achieved on clinical and derived Floodlight features. }
    \label{fig-results}
\end{figure}

Our evaluation focused on the clinically relevant task of predicting 48-week (T48) and 72-week (T72) cCDP using only the initial 12 weeks of data. Overall performance was modest, reflecting the difficulty of long-term prediction from early data (see \cref{fig-results} for T72 results; T48 trends were similar). The highest AUROC achieved with 12 weeks of input was 0.63 (for T72) and 0.61 (for T48), obtained by a transformer using unimodal Moment embeddings, showcasing the potential of FM-derived representations in Floodlight data. 

Despite the challenge of early prediction, we observed encouraging and consistent relative performance differences (\cref{fig-results}). A benefit from multimodality was seen in the transformer architecture: for T72, the multimodal transformer (AUROC 0.63) improved upon a comparable unimodal transformer using only GP-processed Floodlight data (AUROC 0.60), a relative gain suggesting value in combining data sources. A similar advantage was seen for T48 (Multimodal 0.60 vs Unimodal 0.56). 

% FMs demonstrated clear utility over standard approaches. 
FMs demonstrated clear utility over standard machine and deep learning approaches in our disability progression prediction task. For T72, the transformer on Moment embeddings (Unimodal FM, AUROC 0.63) and TabPFN (Unimodal FM, AUROC 0.58) generally outperformed simpler non-FM baselines such as the best sktime ensemble (AUROC 0.52) and standard ML benchmarks (AUROC 0.55). Similar trends held for T48. The evaluated fusion strategies within the custom transformer, however, did not yield substantial gains over the best unimodal FM approach. 

\section{Discussion}

This study demonstrates the potential of foundation models (FMs) and multimodal strategies for the challenging early prediction of MS disability progression. Our modest overall AUROC (0.61-0.63) using 12 weeks of input underscores the difficulty of long-term forecasting from limited early data, yet relative performance differences highlight the value of advanced modeling.

Our findings show FM-derived representations, particularly Moment embeddings with a transformer, effectively capture predictive signals from noisy digital biomarkers, outperforming other FMs on combined features and standard time-series methods. This underscores the importance of tailoring foundation models to the unique properties of different data modalities for effective signal extraction.

Integrating multimodal data also proved beneficial, though challenging. Our custom multimodal transformer (clinical + digital) outperformed its unimodal digital-only counterpart, demonstrating that clinical context enhances predictive signal even when early signals are weak. However, this multimodal approach did not surpass the best-performing unimodal FM (Moment + Transformer). This suggests that the primary difficulty lies not just in the fusion architecture but in the inherent signal-to-noise ratio and heterogeneity of early data, where sparse clinical signals may be underdeveloped and dense digital signals noisy.

From a clinical perspective, achieving an AUROC of 0.63 with only 12 weeks of data, is a promising step for earlier risk stratification. Although not yet enabling high-confidence individual predictions, this benchmark demonstrates the utility of FMs and multimodal learning to improve early risk identification compared to traditional approaches. Continued advancements in robust representation learning from noisy digital data and self-supervised pre-training could further enhance clinical utility.

Related work has shown similar performance for similar tasks, with Mostafa et al., reporting an AUC of 0.63 to 0.67 for predicting only upper limb disability progression in MS \cite{mostafa_2021}. Other works that report higher performance in predicting future disability employ more detailed and harder to retrieve data like imaging (MRI and PET scans) \cite{pinto_prediction_2020}, cerebrospinal fluid and blood biomarkers \cite{comi_2024}, electronic health records in addition to clinical markers \cite{de_brouwer_machine-learning-based_2024}, or passive monitoring data (e.g., sleep quality, fatigue) \cite{chikersal_2022}.

Limitations include reliance on the single CONSONANCE trial, a sample size (N=415) that may constrain highly complex models, and limited modalities (lacking imaging or genomics). Future research may prioritize validation on larger, multi-center datasets, incorporate broader biological modalities, and continue developing robust fusion techniques. Exploring self-supervised pre-training on larger mixed clinical/digital health data repositories before task-specific fine-tuning is a key direction for advancing multimodal FMs in life sciences.
The benchmarking and findings presented have implications beyond MS. Integrating sparse clinical assessments with dense, noisy sensor data is a common challenge in managing chronic diseases via remote patient monitoring. Thus, our insights on FM and multimodal architecture utility could generalize, improving early risk detection and personalized interventions across various medical conditions.

\section{Conclusion}

In this study, we addressed the challenging task of predicting MS disability progression using only 12 weeks of multimodal data. We demonstrate that foundation model embeddings derived from digital biomarkers (Moment embeddings and a transformer model) yield the best predictive performance (AUROC 0.63) among the tested methods, outperforming standard ML and time-series approaches. Furthermore, integrating clinical context via a custom multimodal transformer architecture consistently improved upon its unimodal counterpart, confirming the value of multimodal data even in this difficult early prediction setting. While the inherent limitations of early, heterogeneous data capped overall performance and prevented simple fusion from surpassing the best unimodal FM, our findings clearly show that FMs and multimodal approaches, with potential future extensions towards LLM-based architectures (by treating disparate data types as tokens in a unified sequence, mirroring LLM tokenization strategies), can extract valuable predictive signals beyond traditional methods. These results underscore the potential of applying advanced models to complex clinical prediction tasks and provide a foundation for future research focused on enhancing signal extraction, representation learning, and fusion techniques for challenging real-world health data, ultimately moving towards more personalized and timely patient care.

% \section*{Ethical Statement}

% There are no ethical issues.

% \section*{Acknowledgments}

% Acknowledgments
\clearpage
\onecolumn %Set appendix to one-column format

\appendix
\section*{Supplemental Material}\label{supplement}
\subsection*{Data and Feature Overview} % No number, no entry in TOC

\begin{table*}[h!] % Use table* for full page width in a two-column document. Use table for single column.
\centering
\caption{Data and Feature Overview}
\label{tab:data_overview}
\begin{tabularx}{\textwidth}{|>{\hsize=1\hsize}X|>{\hsize=.5\hsize}X|>{\hsize=1.5\hsize}X|} 
\toprule % Top rule from booktabs
\textbf{Feature} & \textbf{Category} & \textbf{Description} \\
\midrule % Middle rule from booktabs
AGE & Demographic & Patient's age. \\
BBMI & Demographic & Baseline Body Mass Index (BMI). \\
BLSDMT & Clinical History & Baseline Symbol Digit Modalities Test (SDMT) score \cite{smith_symbol_1973}.\\
Average Time Taken for 25-Foot Walk Test & Standard Clinical Assessment & Average time taken to complete a 25-foot walk, assessing ambulation \cite{cutter_1999}. \\
Average of 9 Hole Peg Tests & Standard Clinical Assessment & Average time to complete the 9-Hole Peg Test, assessing manual dexterity \cite{mathiowetz_adult_1985}. \\
Ambulation Score & Standard Clinical Assessment & Assesses patient's walking ability \cite{hauser_ambulation_1983}. \\
Bowel/Bladder Functional System Score & Standard Clinical Assessment & Assesses functional impairment in bowel and bladder systems \cite{kurtzke_rating_1983}. \\
Brainstem Functional System Score & Standard Clinical Assessment & Assesses functional impairment in the brainstem \cite{kurtzke_rating_1983}. \\
Cerebellar Functional System Score & Standard Clinical Assessment & Assesses functional impairment in the cerebellar system \cite{kurtzke_rating_1983}. \\
Cerebral Functional System Score & Standard Clinical Assessment & Assesses functional impairment in the cerebral system \cite{kurtzke_rating_1983}. \\
Expanded Disability Score (EDSS) & Standard Clinical Assessment & Standard measure of MS disability \cite{kurtzke_rating_1983}. \\
NUMRLP & Clinical History & Number of documented relapses. \\
ONSETYRS & Clinical History & Years since the onset of MS symptoms. \\
Pyramidal Functional System Score & Standard Clinical Assessment & Assesses functional impairment in the pyramidal system \cite{kurtzke_rating_1983}. \\
SEX & Demographic & Patient's sex. \\
Sensory Functional System Score & Standard Clinical Assessment & Assesses functional impairment in the sensory system \cite{kurtzke_rating_1983}. \\
Visual Functional System Score & Standard Clinical Assessment & Assesses functional impairment in the visual system \cite{kurtzke_rating_1983}. \\
Max 9HPT & Derived Clinical & Maximum score from the 9-Hole Peg Test. \\
Min 9HPT & Derived Clinical & Minimum score from the 9-Hole Peg Test. \\
Range 9HPT & Derived Clinical & Range of scores from the 9-Hole Peg Test. \\
Cerebellar White Matter Volume & Neuroimaging & Volume of white matter in the cerebellum from MRI. \\
Cerebral White Matter Volume & Neuroimaging & Volume of white matter in the cerebrum from MRI. \\
T2 Lesion Volume & Neuroimaging & Total volume of T2 hyperintense lesions from MRI. \\
Thalamic Volume & Neuroimaging & Volume of the thalamus from MRI. \\
Median step duration & Floodlight (Gait) & Median value of step duration from digital gait measurements \cite{rodrigues_2024}. \\
Median step impulse & Floodlight (Gait) & Median value of step impulse/intensity from digital gait measurements \cite{rinderknecht_2023}. \\
Median step length & Floodlight (Gait) & Median value of step length from digital gait measurements \cite{rodrigues_2024}. \\
Sum of step length & Floodlight (Gait) & Sum of step lengths from digital gait measurements \cite{rodrigues_2024}. \\
Median step velocity & Floodlight (Gait) & Median value of step velocity from digital gait measurements \cite{rodrigues_2024}. \\
Median turn speed & Floodlight (Gait) & Median absolute turn speed from digital gait measurements \cite{cheng_2021}. \\
Trace accuracy for DaS FIGURE 8 shape & Floodlight (Upper limb) & Aggregated tracing accuracy across attempts for the Figure-8 drawing task \cite{graves_preliminary_2023}. \\
Double touching asynchrony in Pinching Test & Floodlight (Upper limb) & Mean time between the first and second finger touch during a pinching test \cite{graves_assessment_2023}. \\
Total number of pinches & Floodlight (Upper limb) & Total number of pinching attempts during a pinching test \cite{graves_assessment_2023}. \\
\bottomrule % Bottom rule from booktabs
\end{tabularx}
\end{table*}

\clearpage
\twocolumn
%% The file named.bst is a bibliography style file for BibTeX 0.99c
\bibliographystyle{named}
\bibliography{ijcai25}

\end{document}